\PassOptionsToPackage{breaklinks=true}{hyperref}
\documentclass[3p,11pt,authoryear]{elsarticle}

\usepackage{fullpage}
\usepackage{xfrac} 
\usepackage{verbatim}
\usepackage[small,bf]{caption}
\usepackage{array}
\usepackage{longtable}
\usepackage{amsbsy}
\usepackage{bbm}
\usepackage{natbib}

\usepackage[utf8]{inputenc} 
\usepackage[T1]{fontenc}    
\usepackage[hidelinks]{hyperref}       
\usepackage{url}            
\usepackage{booktabs}       
\usepackage{amsfonts}       
\usepackage{nicefrac}       
\usepackage{microtype}      
\usepackage{etoolbox}
\usepackage{graphicx}
\usepackage{fullpage}
\usepackage{subcaption}
\usepackage{comment} 

\usepackage{pict2e}

\usepackage{microtype}

\usepackage[dvipsnames]{xcolor}
\usepackage{caption}
\usepackage{colortbl}

\usepackage{mathtools}
\usepackage{times}
\usepackage{latexsym}
\usepackage{multirow}
\usepackage{tabularx}
\usepackage[most]{tcolorbox}
\newtcolorbox{conversationbox}{sharp corners, colback=gray!10, colframe=gray!60, boxrule=0.1mm, boxsep=0mm, breakable}
\newtcolorbox{humaninput}{colback=blue!5, boxrule=0mm, boxsep=1mm, left=1mm, right=1mm, breakable, before upper=\strut, top=1mm, bottom=1mm}
\newtcolorbox{gptoutput}{colback=green!5, boxrule=0mm, boxsep=1mm, left=1mm, right=1mm, breakable, before upper=\strut, top=1mm, bottom=1mm}
\newtcolorbox{claude3output}{colback=yellow!5, boxrule=0mm, boxsep=1mm, left=1mm, right=1mm, breakable, before upper=\strut, top=1mm, bottom=1mm}
\newtcolorbox{geminioutput}{colback=brown!10, boxrule=0mm, boxsep=1mm, left=1mm, right=1mm, breakable, before upper=\strut, top=1mm, bottom=1mm}

\makeatletter
\def\BState{\State\hskip-\ALG@thistlm}
\makeatother

\journal{Transportation Research Part C}

\begin{document}
\begin{frontmatter}

\title{Benchmarking the Capabilities of Large Language Models in Transportation System Engineering: Accuracy, Consistency, and Reasoning Behaviors}


\author[lab1]{Usman Syed}
\author[lab2]{Ethan Light}
\author[lab1]{Xingang Guo}
\author[lab1]{Huan Zhang}
\author[lab3]{Lianhui Qin}
\author[lab2]{ Yanfeng Ouyang}
\author[lab1]{Bin Hu}

\address[lab1]{Electrical \& Computer Engineering, University of Illinois Urbana-Champaign, Urbana, IL 61801, USA}
\address[lab2]{Civil \& Environmental Engineering, University of Illinois Urbana-Champaign , Urbana, IL 61801, USA}
\address[lab3]{Computer Science and Engineering, University of California San Diego, CA,USA}

\begin{abstract}

In this paper, we explore the capabilities of state-of-the-art large language models (LLMs) such as GPT-4, GPT-4o, Claude 3.5 Sonnet, Claude 3 Opus, Gemini 1.5 Pro, Llama 3, and Llama 3.1 in solving some selected undergraduate-level transportation engineering problems.  
We introduce TransportBench, a benchmark dataset that includes a sample of transportation engineering problems on a wide range of subjects in the context of planning, design, management, and control of transportation systems. This dataset is used by human experts to evaluate the capabilities of various commercial and open-sourced LLMs, especially their accuracy, consistency, and reasoning behaviors, in solving transportation engineering problems.  
Our comprehensive analysis uncovers the unique strengths and limitations of each LLM, e.g. our analysis shows the impressive accuracy and some unexpected inconsistent behaviors of Claude 3.5 Sonnet in solving TransportBench problems.  
Our study marks a thrilling first step toward harnessing artificial general intelligence for complex transportation challenges.

\end{abstract}


\end{frontmatter}

\section{Introduction}

In recent years, the advent of artificial intelligence has heralded a new era of computational technology, fundamentally reshaping how we approach problem-solving across diverse domains. Among these, large language models (LLMs) stand out for their sophisticated ability to understand and generate human-like text, offering unprecedented opportunities for innovation across various domains including coding \citep{nijkamp2022codegen, nam2024using, xu2022systematic, chew2023llm, macneil2022generating}, reasoning \citep{wei2022chain, huang2022towards,zhou2022least,sun2023survey, havrilla2024glore}, planning \citep{valmeekam2022large, valmeekam2024planning, zhao2024large, song2023llm, dagan2023dynamic}, mathematics \citep{imani2023mathprompter, azerbayev2023llemma,frieder2024mathematical, zhang2024mathverse, he2023solving}, and science \citep{wang2023scibench,birhane2023science,ouyang2023structured, yeadon2024impact, chen2023bioinfo}.
The scientific community, as well as the general public, have been wondering how soon, if not already, LLMs will be capable of solving complex problems that involve not only general common knowledge, but also advanced, domain-specific terminologies and analytical skills. The educators and researchers are also particularly intrigued by (i) how the advancement of LLMs will impact the future of research, education, and workforce development, and (ii) how the current practice should be adapted or tailored to best accommodate and take full advantage of the LLMs’ capabilities. To answer these questions, the first step is to keep track of and understand the development trends of LLMs with respect to the respective fields of interest, and to identify ways to help LLMs best achieve their positive potential for societal impacts.   
 
Transportation systems engineering is a critical interdisciplinary subfield of civil engineering, focusing on developing principles for the planning, design, operations, and management of all modes of transportation for people and goods \citep{cascetta2009transportation, fricker2004fundamentals}. It encompasses various topics such as transportation economics, driver and vehicle characteristics, guideway geometric design, traffic flow and control, planning and demand modeling, utility and modal split, network analysis, and public transit systems \citep{daganzo2019public}, and is currently undergoing rapid technology-driven revolutions (such as autonomy, connectivity, electrification, and shared economy). The problems in this field combine mathematical foundation with strong engineering, social, and economic principles, making this field particularly ripe for technological interventions with LLMs, and at the same time, an ideal benchmark context for assessing the reasoning capabilities of LLMs.

With such a motivation, our paper studies how state-of-the-art LLMs such as GPT-4 \citep{achiam2023gpt}, GPT-4o \citep{GPT4o}, Claude 3.5 Sonnet \citep{Claude_35}, Claude 3 Opus \citep{Claude_3}, Gemini 1.5 Pro \citep{team2023gemini,geminiteam2024gemini15}, Llama 3 \citep{llama3modelcard}, and Llama 3.1 \citep{llama3modelcard1} can be leveraged to tackle undergraduate-level transportation engineering problems, potentially transforming the landscape of this vital engineering discipline. Understanding such capabilities of LLMs could signify a substantial leap towards a likely and exciting future of transportation engineering in which artificial general intelligence is seamlessly equipped with specialized domain-specific human expertise to enhance productivity and creativity of this engineering field. 
Our paper first introduces TransportBench, a benchmark dataset specifically designed to encapsulate the essential elements of transportation engineering.  
Our TransportBench dataset is designed to cover a wide range of subjects,
form the foundation of our investigation, and enable a structured evaluation of how various LLMs perform in a domain traditionally dominated by human expertise. TransportBench captures the complexities and nuances of transportation engineering problems, providing a rigorous testing ground for assessing the problem-solving abilities of various commercially available and open-sourced LLMs. 
We present evaluations conducted by human experts to evaluate the outputs of these LLM models, focusing on their accuracy and consistency. These evaluations are critical, as they provide insights not only into the effectiveness of each model but also into their potential to integrate with existing engineering practices. 
 Our study also identifies the unique strengths and limitations of each LLM, e.g. our analysis shows both the impressive overall accuracy and some unexpected inconsistent behaviors of Claude 3.5 Sonnet when tested on TransportBench. 
Our contributions can be summarized as follows.

\begin{itemize}
    \item  We introduce an open-source new natural-language dataset called TransportBench, designed to test the capabilities of LLMs in solving undergraduate transportation system problems.
    \item We evaluate the accuracy of GPT-4, GPT-4o, Claude 3 Opus, Claude 3.5 Sonnet, Gemini 1.5 Pro, Llama 3, and Llama 3.1 on TransportBench, conducted by human experts. Our analysis shows that leading LLMs can achieve promising accuracy on TransportBench, and Claude 3.5 Sonnet achieves the highest accuracy on TransportBench among all the LLMs.
    \item We further evaluate the consistency of these LLMs on TransportBench. Interestingly, we observe that Claude 3.5 Sonnet and Claude 3 Opus give less consistent answers when asked to double check their own solutions – this suggests a lack of deep conceptual understanding. In contrast, GPT-4 and GPT-4o give the most consistent responses in such a setting.
    \item We carefully examine the reasoning behaviors of LLMs on TransportBench by twisting some problems from TransportBench. This provides the first study of LLM reasoning in the context of solving basic transportation system problems.   
\end{itemize}

Our study marks a significant step toward harnessing LLMs in the realm of civil engineering. Inspired by the capabilities of LLMs in transportation engineering, it is exciting to envision a promising future where transportation systems are dynamically managed, with AI predicting and mitigating traffic congestion, optimizing maintenance schedules, and even designing next-generation infrastructure with unparalleled efficiency. 
It is our hope that our exploration into the intersection of LLMs and transportation system engineering will set the stage for further research and development, potentially revolutionizing the way we design, build, and maintain our transportation systems in the future.

{\bf Related Work:} The relationship between LLMs and transportation engineering only emerged in the past year or so, but have been discussed in many exploratory papers. 
The relevance of LLMs in the realm of Intelligent Transportation Systems (ITS) has been discussed in \cite{shoaib2023survey}, emphasizing their integral role in advancing transportation intelligence, optimizing traffic management, and designing smart cities. More detailed study on using LLMs for traffic management can be found in \cite{zhang2024trafficgpt}. Discussions on potential applications of LLMs in other ITS problems such as traffic flow prediction, vehicle detection, road condition monitoring, traffic sign recognition, and autonomous
vehicles can be found in \cite{khalil2024advanced}.
The potential of leveraging (and finetuning) ChatGPT in smarter traffic safety decision-making and crash narrative analysis has been discussed in \cite{zheng2023chatgpt, zheng2023trafficsafetygpt,mumtarin2023large}. 
Specifically, TrafficSafetyGPT, finetuned from Llama2 on traffic safety data, is one of the earliest efforts in finetuning LLMs for transportation engineering.  There is also a body of literature on using LLMs for mobility analysis and forecasting \citep{zhang2024large}.   
The potential use of LLMs in interpretation and reasoning tasks related to self-driving has been studied in \cite{cui2024receive}. 
Despite these promising developments on specific topics, there still lacks a comprehensive benchmark study on the capabilities of LLMs in solving basic transportation system problems. 
Our work complements the existing papers by providing such a benchmark study. We conduct a broader and more thorough evaluation of the strengths/weaknesses of leading LLMs such as GPT-4, GPT-4o, Claude 3.5 Sonnet, Claude 3 Opus, Gemini 1.5 Pro, Llama 3, and Llama 3.1 in the context of transportation engineering.

\section{The TransportBench Dataset}

We first create a collection of 140 undergraduate problems that span a broad spectrum of topics including transportation economics, driver characteristics, vehicle motion, road geometry design, traffic flow/control, transportation planning, utility/modal split, transportation networks, and public transit systems. 
TransportBench consists of both the true or false problems and the general Q\&A problems.  We summarize the statistics of our TransportBench dataset for each topic in Table~\ref{tab:controlProblems}.

\begin{table}[t!]
\centering
\caption{Summary of the TransportBench dataset. We report the number of True or False problems, the number of general Q\&A problems, and the total number of problems under each topic. }
\label{tab:controlProblems}
\begin{tabular}{l|ccc}
\toprule
\textbf{Topic}   & \textbf{\# of T or F Prob.}  & \textbf{\# of General Q\&A Prob.} & \textbf{\# of Total Prob.} \\ 
\midrule
Facts   &  14  & 8 & 22 \\
Transportation economics  &0  & 5 & 5  \\
Driver characteristics  & 2  & 3  & 5 \\
Vehicle motion   & 7  & 4 & 11 \\
Geometry design   & 7 & 3 & 10 \\
Traffic flow/control & 8 & 7  & 15 \\
Transportation planning  & 4 & 4 & 8 \\
Utility and modal split  & 4 & 2 & 6\\
Transportation networks  & 0  & 3 & 3  \\
Public transit systems & 27  & 28 & 55   \\
\midrule 
Total  & 73 & 68 & 140  \\
\bottomrule
\end{tabular}
\label{tab1}
\end{table}

\paragraph{Human expertise in data creation} All the problems in TransportBench are selected by the sixth author of this paper based on two of his courses at the University of Illinois: 
a junior-level introductory course CEE 310 - Transportation Engineering (taught 14 times) and a senior-level focused course CEE 418 - Public Transportation Systems (taught 11 times). 
Those for the first nine topics in Table \ref{tab1} are twisted from CEE 310, while those for the last topic ``public transit systems" are from CEE 418. Some of the CEE 418 problems are twisted from a co-authored textbook \citep{daganzo2019public}. Any figure(s) in an original problem are replaced by language descriptions. Together, these problems capture many key areas of current transportation systems engineering, and provide a reasonable benchmark test for LLMs. The solutions to most TransportBench problems are prepared by the second author.

\paragraph{Data format} 
We collect each problem from original documents in PDF files and presentation slides. We manually transfer these problems into LaTeX format. All the problems are carefully verified by human annotators to ensure that LaTeX
documents can be compiled without any syntax errors. In addition, we also provide a detailed step-by-step solution for each problem in LaTeX. For ease of evaluations, we have also provided JSON-formatted files for TransportBench. All the files for our TransportBench dataset are  available  at \url{https://agi4engineering.github.io/TransportBench/}.

\section{Evaluating Accuracy of Leading LLMs on TransportBench}

Now we are ready to evaluate the accuracy of leading LLMs such as GPT-4, GPT-4o,  Claude 3 Opus, Claude 3.5 Sonnet, Gemini 1.5 Pro, Llama 3 (70B) and Llama 3.1 (405B) on TransportBench. 
In this section, we consider a zero-shot prompting strategy (i.e., directly inputting the question) and examine the web-version LLM responses through human expert annotation (our analysis does not require using APIs or local models). The inputs to the web-version LLMs are just the original problem descriptions from TransportBench. 
The zero-shot setting is fundamentally important \citep{kojima2022large} and serves as a meaningful starting point for our analysis.
We first present two examples (i.e. one general Q\&A problem and one True or False problem) to illustrate the zero-shot setting and showcase the capabilities of LLMs in solving transportation problems. First, consider the following transportation economics~problem.
\begin{conversationbox}
\centering{{\bf Zero-Shot Setting (General Q\&A Problem): Problem 2.5 from TransportBench}}
    \begin{humaninput}
    \textbf{Input}: Suppose you have one widget plant at the center of a sufficiently large country of size $S$ mi$^2$, where population is uniformly distributed with a density of $\delta$ people/mi$^2$. The production cost is \$$C$ per unit, and the selling price is \$$P $ per unit, and it can sell $D$ widgets per person per year everywhere in the country. Transportation costs include a fixed cost of $F$ per unit and a variable cost of \$$f$ per widget-mile. Suppose $P > C+F$. Determine the shape, size and the total demand of your market.
    \end{humaninput}
    \begin{gptoutput}
    \textbf{GPT-4}: To solve this problem, we need to determine the region around the widget plant where it is profitable to sell widgets ...
    Since \(P > C + F\), we can determine the break-even distance \(d\) where the selling price equals the total cost: \( P = C + F + fd \quad \implies  d = \frac{P - C - F}{f} \)$\cdots$
The size (area) of the market is
\(  \pi \left( \frac{P - C - F}{f} \right)^2 \). The total market demand is \(\delta \cdot \pi \left(\frac{P - C - F}{f}\right)^2 \cdot D\)
    \end{gptoutput}
     \begin{claude3output}
   \textbf{Claude 3.5 Sonnet}: Let's approach this problem \ldots
Let  P be the revenue, r is the radius of the circular market area, and the total cost $= C + F + fr$ . For the market boundary, revenue equals total cost implies \(  r = (P - C - F)/f \).
Therefore, the market is a circular area with radius $r = (P - C - F)/f$, and the market demand is $\pi \left(\frac{P - C - F}{f}\right)^2 \times \delta \times D$ widgets per year.
    \end{claude3output}
\begin{geminioutput}
     \textbf{Gemini 1.5 Pro}: Here's how we can determine the shape, size, and total demand of the market under the given conditions: ... Since each person buys $D$ widgets per year, the total demand is: $\mbox{Total Demand} = D * \mbox{Population} = D\delta \pi[(P - C - F) / f]^2$. 
\end{geminioutput}
\end{conversationbox}
\noindent For illustrations, the solutions of GPT-4, Claude 3.5 Sonnet, and Gemini  1.5 Pro are provided for the above question. All three LLMs have clearly identified the ``uniformly distributed population" setting in the problem statement and subsequently assumes the use of Euclidean metric. Furthermore, all the LLMs correctly model the profitability of the plant and come up the correct size of the market for the widget plant.
For the second example, we consider a True or False problem on geometry~design.

\begin{conversationbox}
    \centering{{\bf Zero-Shot Setting (True or False Problem): Problem 5.1 from TransportBench}}
    \begin{humaninput}
     \textbf{Input}:  In horizontal curve design, for a given design speed, the larger the road radius, the larger the rate of super-elevation. True or False? 
    \end{humaninput}
     \begin{gptoutput}
    \textbf{GPT-4}:  False. In horizontal curve design, the rate of super-elevation actually decreases as the radius of the curve increases for a given design speed. The rate of super-elevation is used to counteract the lateral acceleration experienced by a vehicle navigating a curve. A larger radius means the curve is less sharp, and therefore, the lateral forces are lower, requiring less super-elevation to maintain vehicle stability and comfort.
    \end{gptoutput}
    \begin{claude3output}
    \textbf{Claude 3.5 Sonnet}: False. In horizontal curve design, for a given design speed, the relationship between road radius and superelevation rate is actually inverse, not directly proportional. Here's a brief explanation:
1. As the radius of a horizontal curve increases, the centrifugal force acting on a vehicle decreases. This means less superelevation is needed to counteract the force. 2. Conversely, tighter curves (smaller radii) require more superelevation to help vehicles safely navigate the turn at the design speed. 3. For a given design speed, as you increase the radius of the curve, you can decrease the rate of superelevation while still maintaining safety and comfort for drivers. This relationship is fundamental to highway geometric design and is typically represented in design charts or tables that show the interplay between curve radius, design speed, and superelevation rate.
    \end{claude3output}
       \begin{geminioutput}
        \textbf{Gemini 1.5 Pro}: False. 
In horizontal curve design, the relationship between road radius and superelevation rate is inverse. This means: 
Larger radius: Requires a smaller rate of superelevation. This is because the centrifugal force experienced by a vehicle traveling at a given speed is lower on a curve with a larger radius.
Smaller radius: Requires a larger rate of superelevation to counteract the increased centrifugal force.
The design goal is to provide a comfortable and safe driving experience by balancing the centrifugal force with the inward tilt of the road surface. 
    \end{geminioutput}
\end{conversationbox}
\noindent In the above example,  all the LLMs have accurately identified the correct relationship between the road radius, super-elevation, and design speed, and come up the correct conclusion that the original statement is false. For concreteness, we listed some detailed reasoning provided by LLMs. We can see that the reasoning from LLMs for this specific question is quite plausible and supports the correct conclusion~well.

Following the above zero-shot setting, we present a comprehensive evaluation of GPT-4, GPT-4o, Claude 3.5 Sonnet, Claude 3 Opus, Gemini 1.5 Pro, Llama 3, and Llama 3.1 on TransportBench. After receiving the LLM responses, we check the correctness of the LLM answers via human annotation. For all the problems, we also check the reasoning provided by LLMs. 
 Our main evaluation metric is Accuracy (\textbf{ACC}), defined as the proportion of instances where the LLMs correctly solve the given problems.  
The ACC of GPT-4, GPT-4o, Claude 3 Opus, Claude 3.5 Sonnet, Gemini 1.5 Pro, Llama 3 (70B), and Llama 3.1 (405B) on the TransportBench dataset is reported in Table~\ref{tab2}.
For simplicity, we only conduct our analysis on one trial per problem\footnote{When we increase the number of trials per problem, the resultant trend is quite similar. See Table \ref{tab:TF_MMRR} in the next section.}, and hence there is some inherent randomness in the ACC obtained. Nevertheless, the results in Table~\ref{tab2} show that leading LLMs such as Claude 3.5 Sonnet, Gemini 1.5 Pro, GPT-4o, Claude 3 Opus, and GPT-4 have shown promising accuracy on TransportBench. 
A few key observations are made below.

\begin{table*}[t!]
  \centering
  \caption{Accuracy (ACC) of GPT-4, GPT-4o, Claude 3 Opus, Gemini 1.5 Pro, Llama 3 and Llama 3.1 on TransportBench. The best results for each topic are highlighted in bold.}
  \label{tab:main_result1}
  \resizebox{1.0\textwidth}{!}{%
    \begin{tabular}{l|c|c|c|c|c|c|c}
      \toprule
      \textbf{Topics} & \textbf{GPT-4} & \textbf{GPT-4o} & \textbf{Claude 3 Opus} & \textbf{Claude 3.5 Sonnet} & \textbf{Gemini 1.5 Pro} & \textbf{Llama 3 (70B)} & \textbf{Llama 3.1 (405B)}\\
      \midrule
      Facts                     & 72.7\% (16/22)         & 72.7\% (16/22)         & 68.2\% (15/22)         & \textbf{81.8\% (18/22)} & \textbf{81.8\% (18/22)} & 68.2\% (15/22)  & 68.2\% (15/22)\\
      Transportation economics  & 20.0\% (1/5)           & \textbf{40.0\% (2/5)}  & \textbf{40.0\% (2/5)}  & \textbf{40.0\% (2/5)}   & \textbf{40.0\% (2/5)}   & 20.0\% (1/5)    & \textbf{40.0\% (2/5)}  \\
      Driver characteristics    & \textbf{100.0\% (5/5)} & \textbf{100.0\% (5/5)} & \textbf{100.0\% (5/5)} & \textbf{100.0\% (5/5)}  & 80.0\% (4/5)            & 40.0\% (2/5) & \textbf{100.0\% (5/5)} \\
      Vehicle motion            & 63.6\% (7/11)          & 81.8\% (9/11)          & 72.7\% (8/11)          & \textbf{90.9\% (10/11)} & 81.8\% (9/11)           & 36.4\% (4/11) & 63.6\% (7/11)  \\
      Geometry design           & 60.0\% (6/10)          & \textbf{70.0\% (7/10)} & \textbf{70.0\% (7/10)} & \textbf{ 70.0\% (7/10)} & \textbf{70.0\% (7/10)}  & 60.0\% (6/10)  & 60.0\% (6/10) \\
      Traffic flow/control      & 40.0\% (6/15)          & 46.7\% (7/15)          & 33.3\% (5/15)          & \textbf{ 53.3\% (8/15)} & \textbf{ 53.3\% (8/15)} & 40.0\% (6/15) & 46.7\% (7/15)  \\
      Transportation planning   & 62.5\% (5/8)           & 50.0\% (4/8)           & 50.0\% (4/8)           & 62.5\% (5/8)            & \textbf{100.0\% (8/8)}  & 50.0\% (4/8) & 50.0\% (4/8) \\
      Utility and modal split    & 66.7\% (4/6)           & \textbf{83.3\% (5/6)}  & 66.7\% (4/6)           & 50.0\% (3/6)            & 50.0\% (3/6)            & 50.0\% (3/6)  & 66.7\% (4/6) \\
      Transportation networks   & 33.3\% (1/3)           & 33.3\% (1/3)           & 66.7\% (2/3)           & \textbf{100.0\% (3/3)}  & \textbf{100.0\% (3/3)}  & 33.3\% (1/3)  & \textbf{100.0\% (3/3)}\\
      Transit systems           & 47.3\% (26/55)         & 50.9\% (28/55)         & 56.4\% (31/55)         & \textbf{60\% (33/55)}   &  49.1\% (27/55)         & 30.9\% (17/55) & 43.6\% (24/55)  \\
      \midrule
      Overall                   & 55.0\% (77/140) & 60.0\% (84/140)& 59.3\% (83/140) & \textbf{ 67.1\% (94/140)}& 63.6\% (89/140) &42.1\% (59/140)  & 55.0\% (77/140)  \\
      \bottomrule
    \end{tabular}
    \label{tab2}
  }
\end{table*}

\begin{itemize}
    \item \textbf{Claude 3.5 Sonnet achieves the best ACC for most topics and the entire TransportBench dataset}. This indicates that in the zero-shot setting, Claude 3.5 Sonnet can be considered as the state-of-the-art LLM in solving problems from TransportBench, demonstrating its superior accuracy in both factual and analytical problem-solving tasks compared to other models. Its performance suggests advanced capabilities in handling complex transportation-related queries.
    \item \textbf{Gemini 1.5 Pro, GPT-4o, and Claude 3 Opus all demonstrate competitive performance.}  Despite the fact that Gemini 1.5 Pro, GPT-4o, and Claude 3 Opus fall short of Claude 3.5 Sonnet in terms of overall accuracy, all these LLMs have achieved reasonably good accuracy on TransportBench.  
    The performance gap between Claude 3.5 Sonnet and Gemini 1.5 Pro is actually not that significant. GPT-4o and Claude 3 Opus can also roughly achieve 60\% ACC. We can see that GPT-4o has surpassed GPT-4 on TransportBench. This is consistent with the known fact that GPT-4o has outperformed GPT-4 on many existing benchmarks across various domains.

    \item \textbf{The open-source model Llama 3.1 has reached the level of the commercial model GPT-4}. There is a huge gap between the ACC achieved by Llama 3 and other LLMs. This may be due to its smaller model size compared to the commercial LLMs.  However, as Meta AI further increases the model size from 70B to 405B, Llama 3.1 has finally reached the ACC level of GPT-4 on TransportBench. Since Llama 3.1 is open-source, it provides valuable insights and a starting point for further research and development. For example, one can potentially improve Llama 3.1 through  instruction finetuning and domain adaption \citep{hulora,weifinetuned}.
    
\end{itemize}

Overall, leading LLMs have shown promise in solving basic transportation system problems. Next, we dig into the performance of these LLMs by examining ACC for different  problem topics and types.

\begin{table*}[t!]
  \centering
  \caption{Accuracy (ACC) of LLMs on  CEE 310 vs. CEE 418 vs. True or False problems vs. General Q\&A. The best results for each metric are highlighted in bold.}
  \label{tab:main_result2}
  \begin{tabular}{l|c|c|c|c}
        \midrule
    \textbf{LLMs} & \textbf{CEE 310}    & \textbf{CEE 418} & \textbf{True or False}    & \textbf{General Q\&A}     \\
    \midrule
    {GPT-4}             & 60.0\% (51/85) & 47.3\% (26/55)   & 58.9\% (43/73)                                & 50.7\% (34/67)             \\
    {GPT-4o}            & 65.9\% (56/85) & 50.9\% (28/55)     & 63.0\% (46/73)                            & 56.7\% (38/67)          \\
    {Claude 3 Opus}     & 61.2\% (52/85) & 56.4\% (31/55)  &  60.3\% (44/73)                    & 58.2\% (39/67)            \\ 
    {Claude 3.5 Sonnet} & 71.8\% (61/85) & \textbf{60.0\% (33/55)} & \textbf{  72.6\% (53/73) }                           &  \textbf{71.6\% (48/67)}\\
    {Gemini 1.5 Pro}    & \textbf{72.9\% (62/85)} & 49.1\% (27/55)   & 71.2\% (52/73)  &                   55.2\% (37/67)           \\
    {Llama 3 (70B)}     & 49.4\% (42/85) & 30.9\% (17/55)   & 45.2\% (33/73)                             & 38.8\% (26/67)          \\
    {Llama 3.1 (405B)}  &  62.4\% (53/85)& 43.6\% (24/55)    &         60.3\% (44/73)           &      49.3\% (33/67)     \\
    \bottomrule
  \end{tabular}
\end{table*}

\paragraph{CEE 310 vs. CEE 418.} 
It is interesting to investigate the impact of problem difficulty levels. Here we study the performance gap of the evaluated LLMs on CEE 418 (the last topic on transit systems) and CEE 310 (all other topics). CEE 310 is an introductory course that covers a very broad range of topics in transportation engineering, while CEE 418 is considered as a more advanced and more focused follow-up course (whose prerequisite is CEE 310). The overall ACC for CEE 310 and CEE 418 can be found in Table \ref{tab:main_result2}. As expected, all the LLMs have lower ACC on CEE 418, and higher ACC on CEE 310.
This confirms that CEE 418 is more challenging than CEE 310 for LLMs.
We notice that an LLM that does better on CEE 310 does not necessarily attains higher ACC on CEE 418. For example, GPT-4o perform better than Claude 3 Opus on CEE 310 problems. However, on CEE 418, Claude 3 Opus is above GPT-4o in terms of ACC. 
Interestingly, Gemini 1.5 Pro is doing extremely well on CEE 310 but has a huge performance drop (23.5\% ACC) on CEE 418. In addition, GPT-4, Llama 3.1, and Llama 3 are all struggling with CEE 418 with very low ACC. In general, the questions from CEE 418 are quite challenging for LLMs. 
 As the most recent LLM, Claude 3.5 Sonnet significantly outperforms GPT-4, GPT-4o, Gemini 1.5 Pro, and Llama 3.1 on CEE 418, demonstrating its superior capabilities in handling more advanced topics in transportation engineering. The following example showcases the impressive reasoning capabilities of Claude 3.5 Sonnet on CEE 418 (we note that Claude 3.5 Sonnet is the only model that gives a correct final answer as well as some plausible reasoning).\footnote{One shall note that while the conclusion of ``41\% larger" is correct, the reasoning is not perfect -- the calculation should have considered walking on both the origin and the destination ends of the trip, and the total expected walking distance should have been $S$ and $S/2$ for a worst-case and an average passenger, respectively. This finding motivates us to investigate deeper into LLMs' reasoning behavior in later sections. \label{footnote1042}}

\begin{conversationbox}
    \centering{{\bf Claude 3.5 Sonnet on CEE 418: Problem 10.42 from TransportBench}}
    \begin{humaninput}
      \textbf{Input}:  In a grid transit network, for realistic analysis/design, how does the optimal stop spacing change if we optimize the average passenger travel time instead of the worst-case passenger travel time? Quantify the change.
    \end{humaninput}
    \begin{claude3output}
    \textbf{Claude 3 Sonnet}: To analyze this change in optimal stop spacing, let's consider the key factors and how they differ between optimizing for average travel time versus worst-case travel time. Let's break this down step-by-step: 1. Key factors:...
4. Mathematical approach:
   Let's use a simplified model to quantify the difference. Assume:
   - A linear route with uniform passenger density
   - Walking speed: $v_w$
   - Vehicle speed between stops: $v_v$
   - Time lost per stop (deceleration, boarding, acceleration): $t_s$.  Let the stop spacing be $S$.
5. Worst-case optimization:
   The worst-case walking distance is $S/2$.
   Total travel time $= (S/2)/v_w + L/v_v + (L/S)*t_s$
   Where $L$ is the total trip length.
   Optimizing this gives an optimal stop spacing of:
   $S_{worst} = sqrt(2*v_v*t_s / v_w)$.
6. Average-case optimization:
   The average walking distance is $S/4$.
   Average travel time $= (S/4)/v_w + L/v_v + (L/S)*t_s$.
   Optimizing this gives an optimal stop spacing of:
   $S_{avg} = sqrt(4*v_v*t_s / v_w)$.
7. Quantifying the change:
   $S_{avg} / S_{worst} = sqrt(4/2) = sqrt(2) \approx 1.41$.
This simplified analysis suggests that optimizing for average travel time results in stop spacing that is approximately $41\%$ larger than when optimizing for worst-case travel time.

    \end{claude3output}
\end{conversationbox}

\paragraph{True or False vs. General Q\&A:} Next, we discuss how the problem type affects the LLM performance. 
The TransportBench consists of True or False problems and general Q\&A problems. Intuitively, True or False problems are easier than more general Q\&A problems. Thus it would be interesting to compare the performance of the LLMs on the two categories. We report the ACC of the seven evaluated LLMs for each problem type in  Table \ref{tab:main_result2}. For most LLMs tested by our study,  we can observe the consistently lower ACC scores for general Q\&A problems compared to True or False problems. This implies that for many LLMs, the True or False problems from TransportBench are easier than the general Q\&A problems. 
Gemini 1.5 Pro achieves high ACC for True or False problems, but its performance significantly degrades on general Q\&A problems.
Noticeably, Claude 3.5 Sonnet achieves similar ACC for both general Q\&A problems (71.6\%) and True or False problems (72.6\%). As the most recent commercial LLM, it has shown impressive improvements on general Q\&A problems.

\section{Evaluating Consistency of LLMs on True or False Problems
}
\label{sec:consistency}

Although (zero-shot) accuracy provides a pivotal metric for initial assessments of LLM capabilities, it alone does not provide a true reflection of the intelligence level of LLMs.
In this section, we will study another important aspect, namely the consistency of LLMs \citep{wang2022self,liang2024internal}. Conceptually, consistency refers to uniform, reliable, and logically coherent responses that maintain the same principles and reasoning across different inquiries.
Consistency can be measured in many different ways. In this section, we study the consistency of LLMs 
on the True or False problems in the following two settings. 

\begin{itemize}
    \item Zero-shot consistency: Firstly, we independently test multiple trials of each problem in the zero-shot setting 
and examine whether LLMs will generate the same answer (either all true or all false). This setting is intuitive and can be illustrated by the following example where Claude 3.5 Sonnet generates inconsistent answers for different independent inquiries. The correct answer is false.
\begin{conversationbox}
  \centering{{\bf Inconsistency in the Zero-Shot Setting: Problem 10.11 from TransportBench}}
    \begin{humaninput}
    \textbf{Input}: Sufficiently large buses (infinite capacity) are used to provide shuttle service, and we always optimize dispatch frequency based on the demand. As the temporal demand distribution becomes more uneven (i.e., heterogeneous over time, but no change in total number of passengers), the total (agency + user) cost increases. True or False?
    \end{humaninput}
     \begin{claude3output}
         \textbf{Claude 3.5 Sonnet (Trial 1)}: To answer this question, let's think through it step by step...
   I believe the statement is \textbf{False}. While it's true that a more uneven demand distribution might lead to some inefficiencies, the ability to optimize dispatch frequency and the assumption of infinite capacity buses should allow for maintaining optimal or near-optimal total costs. The increased costs during peak times would likely be offset by decreased costs during off-peak times.
     \end{claude3output}
     
      \begin{claude3output}
         \textbf{Claude 3.5 Sonnet (Trial 2)}: To answer this question, let's think through it step by step:
1) First, let's consider what factors influence the total cost:- Agency costs: primarily related to the number of buses operated - User costs: primarily related to waiting time
2) In an optimized system, the dispatch frequency is adjusted based on demand. This means:   - During high demand periods, buses are dispatched more frequently   - During low demand periods, buses are dispatched less frequently\ldots
6) Overall effect:   - Agency costs might increase slightly due to the need for a larger fleet- User costs are likely to increase due to longer average waiting times
   - The total cost, being the sum of agency and user costs, would therefore increase. 
Conclusion: \textbf{True}. As the temporal demand distribution becomes more uneven, the total (agency + user) cost is likely to increase, primarily due to increased user waiting times during off-peak periods and potentially higher agency costs for maintaining a larger fleet.
     \end{claude3output}
    
\end{conversationbox}
    \item Consistency under self-checking prompts: Secondly, we examine whether LLMs will generate consistent answers and reasoning when they are prompted to double check their original answers. 
\end{itemize}
It has been recognized from the LLM literature that sometimes LLMs  can correct their mistakes if given simple self-checking prompts, such as "\textit{carefully check your solutions}" \citep{huang2022large,kevian2024capabilities}. However, providing such prompts can also cause the LLMs to flip their original correct answers. 
The second self-checking setting studies such consistency issues and 
incorporates some basic interactions between LLMs and human users. We will provide more details for the self-checking setting~later.

 \begin{table}[t!]
    \centering
    \caption{Zero-shot consistency of LLMs}
    \begin{tabular}{l|c c}
    \hline 
    \textbf{LLMs} & \textbf{Aggregate ACC} $\uparrow$ & \textbf{MRR} $\downarrow$\\
    \hline
    {GPT-4}              & 63.8\% & 23.3\% \\
    {GPT-4o}             & 65.5\% & 23.3\% \\
    {Claude 3 Opus}      & 60.8\% & 24.7\%\\
    {Claude 3.5 Sonnet}  & 75.6\% & 8.2\% \\
    {Gemini 1.5 Pro}         & 70.9\% &  26.0\% \\
    {Llama 3 (70B)}      & 49.8\% & 4.1\%\\
     {Llama 3.1 (405B)}  & 60.5\%  & 12.3\%\\
    \hline 
    \end{tabular}
    \label{tab:TF_MMRR}
\end{table}

On zero-shot consistency of LLMs,  
we test all the LLMs via five independent trials of each of the 73 True or False problems in TransportBench, and report two metrics: 
(i) Mixed Response Rate (MRR) in the zero-shot setting, which is the percentage of the True or False problems that received mixed responses (non-identical answers) in any of the five trials; 
and (ii) aggregate ACC, which is the proportion of the trials where LLMs give the correct true or false label over the total $73 \times 5 = 365$ trials.  
The results are reported in Table \ref{tab:TF_MMRR}.
Notice that a low MRR means that LLMs are consistent in generating either correct or incorrect answers in the zero-shot setting. We need to complement this metric by the aggregate accuracy of LLMs. 
It is desirable to have high aggregate ACC and low MRR at the same time. 
  We make the following observations on the zero-shot consistency of LLMs.

\begin{itemize}
    \item Llama 3 achieves the lowest MRR. However, the aggregate ACC for Llama 3 is also the lowest. This means that Llama 3 has strong bias in answering TransportBench questions and consistently generate wrong answers. After taking a closer look at the responses generated by Llama 3, we realize that  Llama 3 reports True for almost 90\% of the total problem trials. This implies that Llama 3 does not master the technical concepts from transportation engineering.
    \item Llama 3.1 also has a low MRR and achieves higher aggregate ACC than Llama 3 (the aggregate ACC of Llama 3.1 is comparable to GPT-4 and GPT-4o). After examining the response from Llama 3.1 more carefully, we realize that Llama 3.1 has only improved the understanding of the materials from CEE 310 in comparison to Llama 3. We further calculate the aggregate ACC of Llama 3.1 for the T/F problems from CEE 310, which is equal to 63\%. In contrast, the aggregate ACC of Llama 3.1 for the T/F problem from CEE 418 is only 56.3\%. Llama 3.1 is performing reasonably well on the T/F problems for CEE 310. However, when answering the T/F problems from CEE 418, Llama 3.1 still tends to give True most of the time. This implies that Llama 3.1 has not mastered the materials from CEE 418 yet. It will be interesting to see whether the next version of the Llama model still has this biased behaviors for CEE 418. 
    \item Claude 3 Sonnet achieves the highest aggregate ACC while maintaining a very low MRR. In this zero-shot setting, Claude 3 Sonnet can be considered as the state-of-the-art model for solving TransportBench problems. 
    \item From the MRR values, we can see that most LLMs including GPT-4, GPT-4o, Claude 3 Opus, and Gemini 1.5 Pro are not very consistent in giving true/false labels in the zero-shot setting. How to improve their consistency remains open research issues, which should be addressed in the future.
    \item The aggregate ACC trend is quite consistent with the original ACC results in Table \ref{tab:main_result2}. Overall, it is fair to claim that LLMs have shown promise in solving transportation system problems. 
\end{itemize}

When LLMs are prompted to self-check their initial answers, their accuracy may or may not improve.    
The following example illustrates the self-checking setting and showcases that sometimes LLMs can correct their own mistakes in such a self-checking setting.

\begin{conversationbox}
    \centering{\bf Correction via Self-checking: Problem 10.12 from TransportBench}
    \begin{humaninput}
        \textbf{Input}: The agency is trying to decide whether to design a hub-and-spoke network or a grid network to serve a city with uniformly distributed passenger O/D demand. As agency investment (total service route length per unit area) approaches infinity, these two types of networks eventually yield the same expected door-to-door passenger travel time. True or False?
    \end{humaninput}
    \begin{claude3output}
         \textbf{Claude 3.5 Sonnet}: Let's think through this step-by-step:
1) First, let's consider what hub-and-spoke and grid networks look like:
   - Hub-and-spoke: Central hub with routes radiating outward like spokes
   - Grid: Parallel and perpendicular routes forming a grid pattern
2) As investment approaches infinity, we can assume:
   - The number of routes (spokes or grid lines) would approach infinity
   - The spacing between routes would approach zero
3) In a hub-and-spoke network with infinite investment:
   - Every point would essentially have a direct connection to the hub
   - From the hub, there would be a direct connection to every other point
4) In a grid network with infinite investment:
   - The grid would become infinitely fine
   - Any two points would be connected by an almost straight line path
5) Door-to-door travel time in both cases:
   - As the network density approaches infinity, the path between any two points approaches a straight line
   - The travel time would approach the time it takes to travel in a straight line between origin and destination
6) Therefore, with infinite investment, both network types would provide essentially direct, straight-line paths between any two points, resulting in the same expected door-to-door travel time.
Conclusion: \textbf{True}.

    \end{claude3output}
    \begin{humaninput}
    \textbf{Input}: Carefully check your solutions.
    \end{humaninput}
    \begin{claude3output}
         \textbf{Claude 3.5 Sonnet}: You're right to ask me to carefully check my solution. Let's review this problem more critically:
1) The key aspects to consider are:
   - Network structure (hub-and-spoke vs. grid)
   - Uniformly distributed passenger O/D (origin/destination) demand
   - Infinite investment (total service route length per unit area approaching infinity)
   - Expected door-to-door passenger travel time
2) Hub-and-spoke network with infinite investment:
   - As investment approaches infinity, the number of spokes would approach infinity
   - However, all trips not beginning or ending at the hub still require a transfer at the hub
   - This transfer time doesn't disappear, even with infinite investment
3) Grid network with infinite investment:
   - As investment approaches infinity, the grid becomes infinitely fine
   - Any two points can be connected by an almost straight line path
   - No transfers are necessary
4) Door-to-door travel time comparison:
   - Grid network: Approaches direct, straight-line travel time
   - Hub-and-spoke network: Approaches straight-line travel time plus transfer time at the hub
5) Critical realization:
   - The transfer time at the hub in the hub-and-spoke network creates a persistent difference in expected door-to-door travel time compared to the grid network.
Upon careful reconsideration, I realize my previous conclusion was incorrect. 
Revised conclusion: \textbf{False}.
As agency investment approaches infinity, the hub-and-spoke network will still require transfers for many trips, resulting in longer expected door-to-door passenger travel times compared to the grid network.
    \end{claude3output}
    
\end{conversationbox}

\noindent In the above example, Claude 3.5 Sonnet is capable of correcting its original mistake and identifying that the key to this question is that the presence of a hub in hub-and-spoke networks imposes circuity that cannot be avoided even with an infinite route length ---  i.e., the majority of passengers must make a transfer at the hub, which often implies detours. Even though it might be better to say ``no detours are necessary" (instead of ``no transfers are necessary") and avoid "straight-line paths" in the discussion of grid networks, 
this example demonstrates the potential benefits of using self-checking prompts.

However, in the self-checking setting, LLMs may also misadjust their original correct answers and start to generate inconsistent true/false labels. This is illustrated by the following example. 

\begin{conversationbox}
    \centering{\bf Inconsistency in the Self-Checking Setting: Problem 6.3 from TransportBench}
    \begin{humaninput}
        \textbf{Input}: In all stationary traffic, time-mean speed equals space-mean speed. True or~False?
    \end{humaninput}
    \begin{claude3output}
         \textbf{Claude 3.5 Sonnet}: \textbf{False}. In stationary traffic conditions, time-mean speed and space-mean speed are generally not equal. Here's a brief explanation: 1. Time-mean speed is the arithmetic average of speeds of vehicles passing a fixed point over a period of time. 2. Space-mean speed is the harmonic mean of speeds of vehicles over a given length of road at an instant in time. 3. In most cases, time-mean speed is higher than space-mean speed. 4. The difference between these two measures tends to increase with greater speed variability among vehicles. 5. Even in stationary traffic (where traffic conditions are stable over time), there can still be variations in individual vehicle speeds, leading to a difference between time-mean and space-mean speeds.
The equality of time-mean and space-mean speeds occurs only in the special case where all vehicles are traveling at exactly the same speed, which is rarely the case in real traffic conditions, even when stationary.
    \end{claude3output}
    \begin{humaninput}
    \textbf{Input}: Carefully check your solutions.
    \end{humaninput}
    \begin{claude3output}
         \textbf{Claude 3.5 Sonnet}: Thank you for asking me to carefully check my solution. Upon review, I realize I need to correct my previous statement. You're right to question this, as my initial response was incorrect. Let me provide the correct answer: \textbf{True}. In stationary traffic, time-mean speed does equal space-mean speed. Here's the explanation: 1. Stationary traffic refers to a condition where the statistical properties of traffic flow do not change over time. 2. In such conditions, the distribution of speeds at a fixed point (used for time-mean speed) would be the same as the distribution of speeds over a section of road at an instant (used for space-mean speed). 3. When the traffic is stationary, the variability that usually causes the difference between time-mean and space-mean speeds is eliminated. 4. Therefore, in this specific case of stationary traffic, the two measures converge and become equal. I apologize for my initial error. This is an important concept in traffic flow theory, and I'm glad you prompted me to double-check. It's a good reminder of the specific conditions under which these two measures can be~equal.
    \end{claude3output}
    
\end{conversationbox}
\noindent In the above example, Claude 3.5 Sonnet obtains the correct final answer initially in the zero-shot setting (despite the flawed statement in step 2; see a related error to Problem 6.13 in Section 5.3), and then
the self-checking prompt makes it change the conclusion to a wrong one.  
Interestingly, Claude 3.5 Sonnet is zero-shot consistent for this problem, i.e. it generates correct final answers for the five independent trials in the zero-shot setting. 
However, the self-checking prompt 
becomes some disturbance 
that causes Claude 3.5 Sonnet to start giving inconsistent behaviors.
Hence zero-shot consistency and the consistency in the self-checking setting are not equivalent. The interactions between LLMs and human users can lead to unexpected inconsistent behaviors that cannot be observed in a zero-shot setting.

In light of the above discussion, we next provide a complementary perspective on consistency of LLMs using 
two metrics. The first metric is self-checking accuracy (denoted as ACC-$\bar{s}$), which quantifies the
instances in which LLMs give correct answers after the self-checking process. Notice that ACC-$\bar{s}$ does not have to be higher than ACC. 
For an LLM which responds very inconsistently to the self-checking prompts, ACC-$\bar{s}$ can actually be lower than the original ACC in the zero-shot setting due to flipping the initial correct answers. 
The second metric is the number of the True or False problems in which the LLMs flip the original correct answers to wrong ones.
The number of incorrect flips can give a more direct illustration on the extent of the inconsistency in the self-checking setting. For a consistent LLM, we ideally want ACC-$\bar{s}$ to be higher than ACC and the number of incorrect flips to be low. Now we present both metrics for GPT-4, GPT-4o, Claude 3 Opus, Claude 3.5 Sonnet, Gemini 1.5 Pro, Llama 3, and Llama 3.1 in
Table \ref{tab:ACCs_vs_ACCsbar}. We make the following observations.

\begin{table*}[t!]
  \centering
  \caption{Accuracy (ACC) and Absolute Self-Checked Accuracy (ACC-$\bar{s}$) of LLMs on the \textbf{True or False problems}. The best results for each metric are highlighted in bold.}
    \label{tab:ACCs_vs_ACCsbar}
\begin{tabular}{@{}  l|ccccc @{} }
\toprule
  \multicolumn{1}{l}{} & \multicolumn{2}{c}{\textbf{Accuracy}}     & \multicolumn{3}{c}{\textbf{Incorrect flips}} \\ \cmidrule(r){2-6} 
\multicolumn{1}{l}{\textbf{LLMs}}   & \multicolumn{1}{c}{ACC $\uparrow$} & \multicolumn{1}{c}{ACC-$\bar{s}  \uparrow$} & \multicolumn{1}{c}{CEE-418 $\downarrow$}     & \multicolumn{1}{c}{CEE-310 $\downarrow$}    & \multicolumn{1}{c}{Total $\downarrow$}    \\ 
\midrule
\multicolumn{1}{l|}{GPT-4}    &  58.9\% (43/73)                     & \multicolumn{1}{l|}{\textbf{68.5\% (50/73)} }     &        \multicolumn{1}{c|}{\textbf{3}}     &       \multicolumn{1}{c|}{ \textbf{1}}    &      \multicolumn{1}{c}{ \textbf{4} }   \\
\multicolumn{1}{l|}{GPT-4o}    &  63.0\% (46/73)                    & \multicolumn{1}{l|}{\textbf{68.5\% (50/73) }}     &        \multicolumn{1}{c|}{2}     &       \multicolumn{1}{c|}{4}    &      \multicolumn{1}{c}{6}    \\  
\multicolumn{1}{l|}{Claude 3 Opus} & 60.3\% (44/73)                & \multicolumn{1}{l|}{49.3\% (36/73)}               &        \multicolumn{1}{c|}{16}    &      \multicolumn{1}{c|}{18}     &     \multicolumn{1}{c}{34}     \\
\multicolumn{1}{l|}{Claude 3.5 Sonnet} & \textbf{72.6 \% (53/73)}  & \multicolumn{1}{l|}{67.1\% (49/73)}               &        \multicolumn{1}{c|}{8}     &       \multicolumn{1}{c|}{8}     &     \multicolumn{1}{c}{16}     \\
\multicolumn{1}{l|}{Gemini 1.5 Pro} & 71.2\% (52/73)               & \multicolumn{1}{l|}{46.6\% (34/73)}               &        \multicolumn{1}{c|}{11}    &      \multicolumn{1}{c|}{20}     &     \multicolumn{1}{c}{31}   \\
\multicolumn{1}{l|}{Llama 3 (70B)} &45.2\% (33/73)                 & \multicolumn{1}{l|}{56.2\% (41/73) }              &        \multicolumn{1}{c|}{11}    &      \multicolumn{1}{c|}{11}     &     \multicolumn{1}{c}{22}     \\
\multicolumn{1}{l|}{Llama 3.1 (405B)}  & 60.3\% (44/73)            & \multicolumn{1}{l|}{54.8\% (40/73)}               &         \multicolumn{1}{c|}{12}    &     \multicolumn{1}{c|}{13}         &   \multicolumn{1}{c}{25}      \\
\bottomrule
\end{tabular}
\end{table*}

\begin{itemize}
    \item GPT-4o and GPT-4 are the only two models whose accuracy is benefited from the self-checking prompts. For all other models including Claude 3.5 Sonnet, the number of incorrect flips is larger than the number of correct flips. The number of incorrect flips for GPT-4 and GPT-4o is very low. Therefore, in the self-checking setting, GPT-4 and GPT-4o are the most consistent.
    \item Given self-checking prompts, Claude 3.5 Sonnet is still more consistent than Claude 3 Opus, Gemini 1.5 Pro, Llama 3, and Llama 3.1, but less consistent than GPT-4 and GPT-4o. Therefore, for transportation engineering tasks involving human interactions, it is unclear whether one should still view Claude 3.5 Sonnet as the state-of-the-art LLM. 
    \item Claude 3 Opus, Gemini 1.5 Pro, Llama 3, and Llama 3.1 all make many incorrect flips. It seems natural for us to question whether these models really master the concepts involving in our True or False problems. 
\end{itemize}

Overall, we can see that evaluating the consistency of LLMs is actually quite subtle. In the zero-shot setting, Claude 3.5 Sonnet is the most consistent model. However, in the self-checking setting, GPT-4 and GPT-4o are both more consistent than Claude 3.5 Sonnet. 
There are many other settings where the consistency of LLMs. can be tested. For instance, one can rephrase each problem in various tricky ways and then test whether LLMs will be able to answer all the variants correctly. Our analysis is just a starting point, and
future efforts are needed to address the consistency of LLMs in a comprehensive~manner.

\section{Reasoning Behaviors of LLMs on TransportBench}

One impressive observation from our study is that LLMs have started to show promise in reasoning for transportation engineering problems, e.g., sometimes they can provide seemingly correct reasons for their True/False labels or solutions to general Q\&A questions. This observation aligns well with the large body of literature discussing the promise of LLM reasoning across various domains \citep{wei2022chain, huang2022towards,zhou2022least,sun2023survey, havrilla2024glore}.
However, alongside the promise of LLM reasoning, there are potential drawbacks and unresolved issues. For instance, recent work has identified potential breakdowns of LLM reasoning on very simple common-sense problems \citep{nezhurina2024alice}. In this section, we perform a case study of LLM reasoning based on examining LLM responses on TransportBench more carefully. We will discuss various interesting observations on the reasoning behaviors of LLMs in answering transportation engineering problems. Importantly, despite the promise, LLMs can make various subtle reasoning errors.

\subsection{Claude 3.5 Sonnet vs. Claude 3 Opus: Reasoning before conclusion matters}

As seen previously, Claude 3.5 Sonnet has improved the accuracy of Claude 3 Opus on TransportBench. In this section, we will examine the role of reasoning in this improvement for the True or False problems. We will start from the following intriguing example.

\begin{conversationbox}
\centering{{\bf Claude 3.5 Sonnet vs. Claude 3 Opus: Problem 6.5 from TransportBench}}
  \begin{humaninput} 
    \textbf{Input}: In traffic, density = 1 / headway. True or False?  
    \end{humaninput}   
    \begin{claude3output}
   \textbf{Claude 3.5 Sonnet}: To answer this question accurately, let's break down the concepts and their relationship: 1. Traffic density: This is typically defined as the number of vehicles per unit length of road.
2. Headway: In traffic engineering, headway usually refers to the time interval between vehicles passing a point on the road.
Given these definitions, the statement "density = 1 / headway" is not precisely correct. Here's why:
1. Units mismatch: Density is typically measured in vehicles per unit length (e.g., vehicles/km), while headway is measured in time (e.g., seconds). The reciprocal of time wouldn't directly give us vehicles per unit length.
2. Relationship: While there is a relationship between density and headway, it's not as simple as a direct reciprocal. The correct relationship involves speed as well.
The more accurate relationship is: Density = Flow / Speed
Where:
- Flow is measured in vehicles per unit time (e.g., vehicles/hour)
- Speed is measured in distance per unit time (e.g., km/hour)
- Resulting in Density as vehicles per unit distance (e.g., vehicles/km).
Headway is related to flow: Flow = 1 / Headway.
Therefore, the statement "In traffic, density = 1 / headway" is \textbf{False}.
    \end{claude3output}
       \begin{claude3output}
    \textbf{Claude 3 Opus}: \textbf{True}. In traffic flow theory, density is defined as the number of vehicles per unit length of the roadway. Headway, on the other hand, is the time or distance between successive vehicles.
Mathematically, density (k) is the inverse of headway (h), expressed as: $k = 1 / h$.
This relationship assumes a uniform headway between vehicles. In reality, headways can vary, and the density is often calculated as an average over a roadway segment.
   \end{claude3output}
    \begin{humaninput} 
    \textbf{Input (Modified)}: In traffic, density = 1 / time headway. True or False?  
    \end{humaninput}   
       \begin{claude3output}
   \textbf{Claude 3 Opus}: \textbf{True}. In traffic engineering, density is defined as the inverse of the average time headway between vehicles. 
Time headway is the time elapsed between the front of a leading vehicle passing a point on the roadway and the front of the following vehicle passing the same point. It's usually measured in seconds.
So the formula is:
Density = 1 / Average Time Headway,
where density is measured in vehicles per unit time (often vehicles per hour) and average time headway is measured in units of time per vehicle (often seconds per vehicle).
This relationship allows traffic engineers to estimate traffic density by measuring the time gaps between vehicles at a specific point on the road.
 \end{claude3output}
   \end{conversationbox}
\noindent At a first glance, it seems that given the original problem statement, Claude 3.5 Sonnet knows the correct relationship between traffic density and headway, while Claude 3 Opus does not. 
In the concurrent transportation literature, ``headway'' by default refers to ``time headway;'' i.e., the time that elapses between the arrival of two consecutive vehicles at a certain location \citep{daganzo1997, HCM2000}. Some earlier transportation documents used the obsolete term ``space headway'' (or ``distance headway''), which refers to the distance between two consecutive vehicles at a certain time, which has been replaced by ``spacing''. It seems that sometimes Claude 3 Opus cannot distinguish between ``space headway'' and ``headway.''
 However, even if we explicitly state ``time headway" in the problem statement, Claude 3 Opus still cannot give the correct conclusion. After taking a closer look at the above example, we notice that Claude 3.5 Sonnet first provided a detailed reasoning process before giving the final conclusion, while Claude 3 Opus first gave a conclusion and then provided some explanations to justify the answer. This motivates us to  make two modifications to the original problem, as shown below. 

\begin{conversationbox}
\centering{{\bf Reasoning Matters: Adding Prompts to Problem 6.5 from TransportBench}}
  \begin{humaninput} 
    \textbf{Input (Modification 1)}: In traffic, density = 1 / headway. True or False?  Provide detailed reasoning before giving the final conclusion.
    \end{humaninput}   
    \begin{claude3output}
   \textbf{Claude 3 Opus}: To determine if the statement "In traffic, density = 1 / headway" is true or false, let's first define the terms and explore their relationship:
Traffic Density: Traffic density is the number of vehicles occupying a given length of a highway or road. It is usually expressed as vehicles per kilometer or vehicles per mile.
Headway: Headway is the time gap between the front of a leading vehicle and the front of the following vehicle, measured at a specific point on the road. It is usually expressed in seconds.
The fundamental relationship between density and headway is: density = flow / speed where:
- density is the traffic density (vehicles/km or vehicles/mile)
- flow is the traffic flow rate (vehicles/hour)
- speed is the average speed of the vehicles (km/hour or miles/hour). Assuming a constant flow rate, the relationship between headway and flow is:
headway = 1 / flow.
Substituting this into the density equation, we get:
density = flow / speed = (1 / headway) / speed.
Therefore, the statement "density = 1 / headway" is \textbf{false}. The correct relationship is:
density = (1 / headway) / speed.
In other words, density is inversely proportional to both headway and speed. As headway decreases (vehicles become closer together), density increases, assuming a constant speed. However, density is not simply the reciprocal of headway, as the statement suggests.
 \end{claude3output}
  \begin{humaninput} 
    \textbf{Input (Modification 2)}: In traffic, density = 1 / headway. True or False?  Explain your answer.
    \end{humaninput}   
    \begin{claude3output}
   \textbf{Claude 3.5 Sonnet}: \textbf{True}. Let me explain why this statement is correct: 1. Definition of terms:
   - Density: The number of vehicles per unit length of road (usually vehicles per kilometer or mile)
   - Headway: The time or distance between two consecutive vehicles in a traffic stream.
2. Relationship:
   - Headway is typically measured in time (seconds) or distance (meters or feet)
   - Density is the inverse of the average space headway.
3. Explanation:
   - If we consider a uniform traffic flow where all vehicles are evenly spaced:
     - A longer headway means fewer vehicles in a given length of road (lower density)
     - A shorter headway means more vehicles in a given length of road (higher density)
   - Mathematically, we can express this as:
     Density = 1 / (Average space headway)
   - Note: When using time headway, you need to factor in the speed to convert it to space headway:
     Density = 1 / (Speed * Average time headway).
4. Example:
   - If the average headway is 50 meters between vehicles:
     Density = 1 / 50 meters = 0.02 vehicles per meter = 20 vehicles per kilometer.
This inverse relationship between density and headway is fundamental in traffic flow theory and is used in various traffic engineering calculations and models.
 \end{claude3output}
   \end{conversationbox}
\noindent 
First, we give the prompt ``provide detailed reasoning before giving the final conclusion" to Claude 3 Opus. All of a sudden, Claude 3 Opus can answer this question correctly. 
It correctly points out that headway is the time gap between the front of a leading vehicle and the front of the following vehicle, measured at a specific point on the road. 
Next, we give the prompt ``explain your answer" to Claude 3.5 Sonnet such that it will first give a True or False label. Now Claude 3.5 Sonnet 
struggles with the concept ``headway" (it defines headway to be the time or distance between two consecutive vehicles in a traffic stream). From these behavior changes, 
it seems that LLMs are more likely to distill the right knowledge when asked to provide detailed reasoning before giving the conclusion. 
This intuitively makes sense with our understanding that LLMs generate their response based on text strings in an autoregressive left-to-right manner. A similar example is shown as follows (the correct answer is False).

\begin{conversationbox}
\centering{{\bf Reasoning matters: Problem 6.7 from TransportBench}}
    \begin{humaninput} 
    \textbf{Input}: In a cumulative diagram for a traffic network, the horizontal separation between the departure and the arrival curves gives the number of vehicles between the two observers.  True or False? 
    \end{humaninput}
    \begin{claude3output}
    \textbf{Claude 3.5 Sonnet Opus}: To answer this question accurately, let's think through it step by step: 1) First, let's  \ldots 
7) The horizontal separation, on the other hand, does not directly represent the number of vehicles. It represents a time difference.
Given this analysis, we can conclude:
\textbf{False}. 
    \end{claude3output}
     \begin{claude3output}
   \textbf{Claude 3 Opus}: \textbf{True}. In a cumulative diagram for a traffic network, the vertical axis represents the cumulative number of vehicles, while the horizontal axis represents time. The diagram typically shows two curves: one for the departure of vehicles at an upstream location and another for the arrival of those vehicles at a downstream location.
The horizontal separation between these two curves at any given point in time represents the number of vehicles that have departed the upstream location but have not yet arrived at the downstream location. In other words, it shows the number of vehicles that are currently traveling between the two observation points at that specific time.
This principle is used in traffic flow analysis to understand the distribution of vehicles along a road segment and to identify potential bottlenecks within the network.
    \end{claude3output}
     \begin{humaninput} 
    \textbf{Input (Modification 1)}: In a cumulative diagram for a traffic network, the horizontal separation between the departure and the arrival curves gives the number of vehicles between the two observers.  True or False? Provide detailed reasoning before giving the final conclusion.
    \end{humaninput}
     \begin{claude3output}
   \textbf{Claude 3 Opus}:
    To determine whether the given statement is true or false, let's first understand what a cumulative diagram is and how it represents traffic flow\ldots
In conclusion, the given statement is \textbf{false}. The horizontal separation between the departure and arrival curves in a cumulative diagram represents travel time, while the vertical separation represents the number of vehicles between the two observation points.
     \end{claude3output}
       \begin{humaninput} 
    \textbf{Input (Modification 2)}: In a cumulative diagram for a traffic network, the horizontal separation between the departure and the arrival curves gives the number of vehicles between the two observers.  True or False? Explain your answer.
    \end{humaninput}
     \begin{claude3output}
   \textbf{Claude 3.5 Sonnet}:
  \textbf{True}. Let me explain why this statement is correct\ldots
     \end{claude3output}
\end{conversationbox}
\noindent The above example again highlights the potential important role of LLM reasoning for generating correct final conclusions. While the prompt ``provide detailed reasoning before giving the final conclusion" is not a universal fix to many LLM errors, the order of reasoning and conclusion generation sometimes matters.

\subsection{Correct final answers may come from flawed reasoning approaches}

The next important observation is that LLMs can make unclear/flawed reasoning to generate correct final conclusions for certain problems, giving  illusions on their proficiency in transportation system topics. We have seen this issue with some of the examples in Section 3.
The following example provides a clear illustration for this.

\begin{conversationbox}
    \centering{{\bf Flawed Reasoning with Correct Conclusion: Problem 10.55 from TransportBench}}
    \begin{humaninput}
      \textbf{Input}:  A bus line runs from 5 am to midnight. In each direction, the line has 40 stops and is 12 km long. The bus cruising speed is 30 km/h, and each stop imposes a delay of 15 seconds. The headway is 10 min during the rush hours (7 am-9 am, and 3 pm-7 pm), and 30 min for the remainder of the service period. What is the minimum fleet size M needed to run this~route? 
    \end{humaninput}
    \begin{claude3output}
    \textbf{Claude 3 Sonnet}: Let's approach this problem step by step:     
    1. Calculate the round-trip time:
   - Distance: 12 km × 2 (round trip) = 24 km
   - Time at cruising speed: 24 km ÷ 30 km/h = 0.8 hours = 48 minutes
   - Number of stops: 40 × 2 (round trip) = 80 stops
   - Total delay at stops: 80 × 15 seconds = 1200 seconds = 20 minutes
   - Total round-trip time: 48 + 20 = 68 minutes.
2. Determine the number of buses needed during different periods:
   a) Rush hours (headway = 10 minutes):
      Buses needed = Round-trip time ÷ Headway
      = 68 ÷ 10 = 6.8, rounded up to 7 buses

   b) Non-rush hours (headway = 30 minutes):
      Buses needed = 68 ÷ 30 = 2.27, rounded up to 3 buses.
3. Calculate the minimum fleet size:
   The minimum fleet size will be the maximum number of buses needed at any time, which is during rush hours.
Therefore, the minimum fleet size M needed to run this route is 7 buses.
    \end{claude3output}
      \begin{humaninput}
      \textbf{Input (Modification 1)}:  A bus line runs from 5 am to midnight. In each direction, the line has 40 stops and is 15 km long. The bus cruising speed is 25 km/h, and each stop imposes a delay of 15 seconds. The dispatch headway from a terminus (at one end of the route) is 10 min during the rush hours (7 am-8 am, and 5 pm-6 pm), and 30 min for the remainder of the service period. What is the minimum fleet size M needed to run this route?
    \end{humaninput}
    \begin{claude3output}
    \textbf{Claude 3 Sonnet}: Let's approach this problem step by step:
1. Calculate the total trip time: Distance = 15 km, Cruising speed = 25 km/h, Travel time = 15 km ÷ 25 km/h = 0.6 hours = 36 minutes, Stop delay = 15 seconds × 40 stops = 600 seconds = 10 minutes. Total one-way trip time = 36 + 10 = 46 minutes.
2. Calculate the round trip time: Round trip time = 46 × 2 = 92 minutes.
3. Determine the number of buses needed for different headways: For 10-minute headway (rush hours): Buses needed = Round trip time ÷ Headway = 92/10 = 9.2, Round up to 10 buses.
For 30-minute headway (non-rush hours): Buses needed = 92 ÷ 30 = 3.07, Round up to 4 buses.
4. Determine the minimum fleet size:
The minimum fleet size will be the larger of the two calculated values, which is 10 buses.
    \end{claude3output}
\end{conversationbox}

\noindent In the above example, we first test LLMs on Problem 10.55 from TransportBench. We can see that Claude 3.5 Sonnet is able to get the correct answer $M=7$. However, upon reading the reasoning behind the answer, we find it to be flawed since it simply computes the ``steady-state" vehicle requirements in peak and off-peak periods separately (and chooses the larger value), but ignores that some of the vehicles dispatched in the off-peak period could later return from service and become available for use during the period periods. In other words, LLMs implicitly assume in their reasoning that rush hour service is run from $-\infty$ to $\infty$ time. 
Such a reasoning error is quite subtle and does not affect the final number of $M=7$ in the original problem, partly because the peak period duration is sufficiently long to offset the impacts of returned vehicles from the off-peak period. When we modify the problem statement by shortening the peak period durations, the fleet size requirement should decrease to $M=8$ (instead of $10$). This time Claude 3.5 Sonnet uses the same reasoning which eventually leads to the wrong answer. This example highlights the fact that sometimes LLMs can only use a simplistic reasoning path without fully understanding the physics behind the transportation engineering problem. 

Another example can be obtained by examining  the general Q\&A form of Problem 10.16 from TransportBench.

\begin{conversationbox}
    \centering{{\bf True or False vs. General Q\&A Versions: Problem 10.16 from TransportBench}}
    \begin{humaninput}
      \textbf{Input (True or False Form)}:   The worst-case waiting time of passengers traveling in a two-level hierarchical corridor transit system (with common headway $H$), when they have an appointment at the destination and know the schedule, is $3H$. True or False?
    \end{humaninput}
    \begin{claude3output}
    \textbf{Claude 3 Sonnet}: To answer this question, let's think through it step by step: 1) In a two-level hierarchical corridor transit system, we typically have:
   - A main line (trunk line)
   - Secondary lines (feeder lines) that connect to the main line. 2) The common headway $H$ means that all lines, both main and secondary, have the same frequency of service.
3) For a passenger to reach their destination, they might need to:
   - Wait for a feeder line
   - Travel on the feeder line
   - Transfer to the main line
   - Travel on the main line.
4) The worst-case scenario would involve:
   - Just missing a feeder line (waiting almost full headway H)
   - Transferring to the main line (potentially waiting another full headway H).
5) However, since the passenger knows the schedule, they can time their arrival at the first stop to minimize waiting time.
6) In the absolute worst case, a passenger might:
   - Wait up to H for the feeder line
   - Just miss the connection to the main line, waiting another H
   - Arrive at their destination, potentially waiting up to H for their appointment time.
7) This worst-case scenario adds up to a total waiting time of 3H.
Therefore, the statement is True.
    \end{claude3output}
      \begin{humaninput}
      \textbf{Input (Q\&A Form)}:  Calculate the worst-case waiting time of passengers traveling in a two-level hierarchical corridor transit system (with common headway $H$), when they have an appointment at the destination and know the schedule.
    \end{humaninput}
    \begin{claude3output}
    \textbf{Claude 3 Sonnet}: To calculate the worst-case waiting time in this scenario, let's break down the problem step by step:
1) First, let's consider what a two-level hierarchical corridor transit system means:
   - There are two levels of service: a main line and feeder lines.
   - Passengers may need to transfer from a feeder line to the main line.
2) Given information:
   - The system operates with a common headway H.
   - Passengers know the schedule and have an appointment at the destination.
3) In the worst-case scenario:
   - A passenger just misses a feeder line vehicle.
   - After taking the feeder line, they just miss the main line vehicle.
   - They arrive at their destination just before their appointment.
4) Worst-case waiting time calculation:
   a) Wait for feeder line: $H$
   b) Travel time on feeder line: (Let's call this $T_f$, but it doesn't affect waiting time)
   c) Wait for main line: $H$
   d) Travel time on main line: (Let's call this $T_m$, but it doesn't affect waiting time)
5) Total worst-case waiting time: $W_{worst} = H + H = 2H$
Therefore, the worst-case waiting time for passengers in this system is $2H$, where $H$ is the common headway.
    \end{claude3output}
\end{conversationbox}

\noindent For the above example, Claude 3.5 Sonnet 
gives the wrong value $3H$ for the worst-case waiting time when the problem is stated as a True or False problem. The correct reasoning shown in step 5) is ignored in steps 6) and 7). Interestingly, if we state the problem in a general Q\&A form and directly ask for calculating the worst-case waiting time, Claude 3.5 Sonnet gives the right answer $2H$. 
However, a closer look at its reasoning shows that the seemingly correct answer actually comes from two simultaneous mistakes: assuming waiting time $H$ for the feeder line (which should not be included because the passengers know the schedule), and ignoring the waiting time $H$ for the appointment (which should be included). Again, this demonstrates that sometimes LLMs can obtain final correct values with flawed~reasoning.

\subsection{Simple domain-specific prompts can improve reasoning}

Finally, we also notice that sometimes simple domain-specific prompts can notably improve LLM reasoning. For illustration, consider the following problem.

\begin{conversationbox}
\centering{{\bf Simple Domain-Specific Prompts: Problem 4.5 from TransportBench}}
    \begin{humaninput} 
    \textbf{Input}: If the speed of a vehicle keeps increasing, the aerodynamic resistance will eventually exceed the rolling resistance, because the aerodynamic resistance depends on the velocity of the vehicle while the rolling resistance does not. True or False? 
    \end{humaninput}
     \begin{claude3output}
   \textbf{Claude 3.5 Sonnet}: To answer this question, let's break down the key concepts and their relationships:
1. Aerodynamic resistance (drag):
   - Depends on the velocity of the vehicle
   - Typically increases quadratically with velocity (proportional to $v^2$) 
2. Rolling resistance:
   - Primarily depends on the weight of the vehicle and road conditions
   - Generally does not depend on velocity (or has a very weak dependence). Given these facts, we can conclude:
\textbf{True}.

    \end{claude3output}
     \begin{humaninput} 
    \textbf{Input (Modified)}: If the speed of a vehicle keeps increasing, the aerodynamic resistance will eventually exceed the rolling resistance, because the aerodynamic resistance depends on the velocity of the vehicle while the rolling resistance does not. True or False? Think about rolling resistance at different speed values.
    \end{humaninput}
      \begin{claude3output}
   \textbf{Claude 3.5 Sonnet}: Let's think about this step-by-step: 1. Aerodynamic resistance:
   - Also known as air resistance or drag
   - Increases with velocity, typically proportional to the square of velocity.
2. Rolling resistance:
   - Caused by the deformation of the wheel and the surface it's rolling on
   - Generally considered to be independent of velocity for low speeds
   - However, at higher speeds, rolling resistance does actually increase.
3. Rolling resistance at different speeds:
   - At low speeds, rolling resistance is relatively constant
   - As speed increases, several factors cause rolling resistance to increase:
     a. Hysteresis in the tire material increases
     b. Centrifugal forces cause tire deformation
     c. Vibrations in the tire structure increase.
4. Relationship between the two resistances:
   - At low speeds, rolling resistance is typically the dominant force
   - As speed increases, aerodynamic resistance increases more rapidly (proportional to $v^2$)
   - Rolling resistance also increases with speed, but not as dramatically as aerodynamic resistance
5. Conclusion:
   While it's true that aerodynamic resistance will eventually exceed rolling resistance as speed increases, the statement as given is not entirely accurate. It's incorrect to say that rolling resistance does not depend on velocity. It does, just not as strongly as aerodynamic resistance.
Therefore, the statement is \textbf{False}. A more accurate statement would be:
``If the speed of a vehicle keeps increasing, the aerodynamic resistance will eventually exceed the rolling resistance, because the aerodynamic resistance increases more rapidly with velocity compared to the rolling~resistance."
    \end{claude3output}

\end{conversationbox}

\noindent In the above example, the problem statement is in the form of "A is true, because B and C are true." Both A and B are true, but C is not --- i.e., the rolling resistance does increase with the velocity: 
it remains almost constant only ``within a range of small speeds," but 
for higher speeds, it grows as the speed increases. 
Claude 3.5 Sonnet  fails in catching this reasoning error in the ``because" statement 
and stated that the rolling resistance is relatively independent of vehicle. 
It appears that  
the LLM either jumps to the wrong conclusion simply because ``A" is true, or  it does not recognize the right condition for ``C" to hold.
Interestingly, once we give a simple domain-specific prompt ``think about rolling resistance at different speed values,"
  Claude 3.5 Sonnet gives the correct answer under the reason that it is incorrect to say that rolling resistance does not depend on velocity.
Below is another relevant example.

\begin{conversationbox}
\centering{{\bf Simple Domain-Specific Prompts: Problem 6.13 from TransportBench}}
    \begin{humaninput} 
    \textbf{Input}:  Consider a two-lane highway with stationary traffic. There are two types of vehicles, cars and trucks, each occupying one lane. Cars travel with constant speed $v_c$ and the time headway between any two consecutive cars is $h_c$. Trucks travel with constant speed $v_t$, and the headway between any two consecutive trucks is $h_t$. 
\begin{enumerate}
    \item[a.] What is the total density (regarding all vehicles) on this highway?
    \item[b.] What is the space-mean speed? 
\end{enumerate}
    \end{humaninput}
    \begin{claude3output}
    \textbf{Claude 3.5 Sonnet}: Let's approach this problem step by step. We'll start by defining some terms and then solve each part of the question. Definitions:
- Density ($k$): The number of vehicles per unit length of the roadway. - Space-mean speed ($v_s$): The harmonic mean of the speeds of vehicles passing a point on the roadway.
Let's solve each part:
a. Total density on the highway:
To find the total density, we need to calculate the density for cars and trucks separately and then add them together.
For cars:
- Density of cars $(k_c) = 1 / (v_c * h_c)$.
For trucks:
- Density of trucks $(k_t) = 1 / (v_t * h_t)$.
Total density $(k) = k_c + k_t
                  = 1 / (v_c * h_c) + 1 / (v_t * h_t)$.
b. Space-mean speed:
The space-mean speed is calculated using the harmonic mean of the speeds, weighted by the proportion of vehicles of each type.
Let's define:
- $f_c$: Fraction of cars $= k_c / (k_c + k_t)$
- $f_t$: Fraction of trucks $ = k_t / (k_c + k_t)$. Space-mean speed $(v_s) = 1 / (f_c/v_c + f_t/v_t)$.
Substituting the fractions:
$v_s = 1 / ((k_c/(k_c+k_t))/v_c + (k_t/(k_c+k_t))/v_t)$.
Simplifying:
$v_s = (k_c + k_t) / (k_c/v_c + k_t/v_t)$.
Substituting the density expressions:
$v_s = (1/(v_c*h_c) + 1/(v_t*h_t)) / (1/(v_c^2*h_c) + 1/(v_t^2*h_t))$.
To summarize:
a. Total density: $k = 1 / (v_c * h_c) + 1 / (v_t * h_t)$
b. Space-mean speed: $v_s = (1/(v_c*h_c) + 1/(v_t*h_t)) / (1/(v_c^2*h_c) + 1/(v_t^2*h_t))$
    \end{claude3output}
  \begin{humaninput} 
    \textbf{Input (Modified)}:  Consider a two-lane highway with stationary traffic. There are two types of vehicles, cars and trucks, each occupying one lane. Cars travel with constant speed $v_c$ and the time headway between any two consecutive cars is $h_c$. Trucks travel with constant speed $v_t$, and the headway between any two consecutive trucks is $h_t$. 
\begin{enumerate}
    \item[a.] What is the total density (regarding all vehicles) on this highway?
    \item[b.] What is the space-mean speed? (Notice that space-mean speed is the  arithmetic mean of vehicle speeds measured from an aerial photo.)
\end{enumerate}
    \end{humaninput}

     \begin{claude3output}
   \textbf{Claude 3.5 Sonnet}: 
Let's approach this problem step by step:
1) First, let's define some variables:
   $k_c$ = density of cars
   $k_t$ = density of trucks
   $k$ = total density of vehicles
2) Now, let's solve each part:
a) To find the total density:
   - For cars: 
     Density = 1 / (speed * headway)
     $k_c = 1 / (v_c * h_c)$
   - For trucks:
     $k_t = 1 / (v_t * h_t)$
   - Total density is the sum of these:
     $k = k_c + k_t = 1 / (v_c * h_c) + 1 / (v_t * h_t)$
b) To find the space-mean speed:
   The space-mean speed ($v_s$) is given by:
   $v_s = \frac{q}{k}$
   Where $q$ is the flow rate and $k$ is the total density.
   - Flow rate for cars: $q_c = k_c * v_c = v_c / (v_c * h_c) = 1 / h_c$
   - Flow rate for trucks: $q_t = k_t * v_t = v_t / (v_t * h_t) = 1 / h_t$
   - Total flow rate: $q = q_c + q_t = 1/h_c + 1/h_t$.
 Now we can calculate the space-mean speed:
 $v_s = \frac{q}{k} = \frac{1/h_c + 1/h_t}{1 / (v_c * h_c) + 1 / (v_t * h_t)}= \frac{v_c * v_t * (h_c + h_t)}{v_t * h_t + v_c * h_c}$.
Therefore, the final answers are:
a) Total density: $k = 1 / (v_c * h_c) + 1 / (v_t * h_t)$
b) Space-mean speed: $v_s = \frac{v_c * v_t * (h_c + h_t)}{v_t * h_t + v_c * h_c}$.
    \end{claude3output}
\end{conversationbox}
\noindent For the above example, Claude 3.5 Sonnet can compute the total vehicle density for a two-lane highway, but reasons incorrectly to compute the space-mean speed. In all of our trials, it repeatedly states that the space-mean speed is ``the harmonic mean of the individual speeds, weighted by the proportion of vehicles of each type." And then it uses the densities of the two vehicle types as the weights. This is wrong, however, since harmonic mean shall be used only when speeds are weighted by the flow of either vehicle type (as if speed samples are collected by a roadside observer); while arithmetic mean should be used directly when speeds are weighted by density (as if speed samples are collected from an aerial video). 
This example shows that Claude 3.5 Sonnet has not fully understood the subtle connection between the space-mean speed formula and the way speed samples are collected. 
In light of this observation, we try to give the domain-specific prompt ``space-mean speed is the arithmetic mean of vehicle speeds measured from an aerial photo." Then, Claude 3.5 Sonnet is able to avoid the ``harmonic mean" + ``weighted by density" trap, and come up the correct reasoning.  

From the above examples, we can see that simple domain-specific prompts may be quite useful in helping LLMs improve their reasoning in the domain of transportation engineering. An interesting future research direction is to design systematic principled prompts for bigger classes of transportation engineering problems.

\section{Conclusion and Future Work}
In this paper, we introduce the TransportBench dataset for the purpose of benchmarking the capabilities of leading LLMs in solving undergraduate-level transportation engineering problems. Our benchmark study 
highlights Claude 3.5 Sonnet as the most proficient model in this domain. Our findings demonstrate the strong potential and promise of using leading LLMs to revolutionize problem solving in the field of transportation engineering. In particular, it appears that current LLMs are quite capable of memorizing facts and technical definitions, connecting related concepts, and integrating simple symbolic derivations into basic logical reasoning --- as a result, they perform relatively well on True/False problems and some of the simpler general Q\&A problems. However, careful assessments by domain experts have also illuminated the limitations of these LLMs, particularly noting the necessity for improved reasoning and explanatory capabilities before they can be used for practical problems in real-world applications. Once the problems require more detailed analysis of the underlying physical processes (e.g., the bus fleet size problem 10.55), or when they require a deeper understanding of subtle concepts or statements (e.g., the space-mean speed problem 6.13(b), or the rolling resistance problem 4.5), LLMs seem to face notable difficulties.  

Looking ahead, we feel that future research on LLM applications in transportation engineering should focus on (i) enhanced  pre-training, finetuning, and evaluations with expanded transportation systems problem datasets, especially with more general Q\&A problems in additional subject areas and at more advanced (senior or graduate) levels; (ii) systematic study of domain-specific prompting and in-context learning for solving more complex transportation engineering tasks, possibly by holistically categorizing basic solution steps or approaches for various transportation problems and passing those high-level information to LLMs; (iii) developing reliable LLM agents with strong tool-use abilities such as leveraging external coding tools or specialized modeling or simulation platforms for sub-domains like geometric design, network analysis, traffic management, and demand modeling; (iv) improving   LLM reasoning for transportation engineering via development and integration of advanced search algorithms (e.g., Tree-of-Thoughts \citep{yao2024tree}, Reasoning-via-Planning \citep{hao2023reasoning}, etc.) with domain knowledge; and (v) exploring the potential use of LLMs in interdisciplinary areas that connect transportation engineering to other related domains such as urban and regional planning, other civil engineering subfields\footnote{For instance, LLMs have already been recently used in areas such as water engineering \citep{xu2024unlocking}.} (e.g., hydrology, construction), and other engineering disciplines (e.g., mechanical engineering, electrical engineering, computer science), and broadly, other sciences (e.g., social science, climatology).
By advancing research in these directions, we can further harness AI's capabilities to help human experts transform the future transportation engineering, ultimately leading to smarter, safer, and more sustainable transportation systems.

\bibliography{mainbib}
\bibliographystyle{plainnat}

\end{document}